\documentclass[landscape,twocolumn,10pt]{article}
 \usepackage[dvips]{graphics,graphicx,epsfig}
\usepackage[]{amssymb,amsmath, graphicx, subfigure, epsfig, multirow}

\def\IR{{\rm I \kern-0.20em R}}
\def\bbbr{{\IR}}
\newtheorem{thm}{Theorem}

\newtheorem{defn}{Definition}
\newtheorem{lem}{Lemma}

\begin{document}
\date{}    
\title{\bf Scalable Algorithms for Aggregating \\Disparate Forecasts of Probability\thanks{This research was supported in part by the Army Research Office under Grant
DAAD19-00-1-0466, in part by the U. S. Army Pantheon Project, in part by Draper Laboratory under Grant IR\&D 6002, and in part by the National Science Foundation under Grants CCR-0020524, 
CCR-0312413 and IIS-9978135.}}
\author{
J. B. Predd\,\,\,\,\,\,\,\,\,\,\,\,S. R. Kulkarni\,\,\,\,\,\,\,\,\,\,\,\,H. V. Poor\\
Department of Electrical Engineering\\
Princeton University\\
Princeton, NJ 08544\\
$\{$jpredd,kulkarni,poor$\}$@princeton.edu
\and
D. N. Osherson\\
Department of Psychology\\
Princeton University\\
Princeton, NJ 08544\\
osherson@princeton.edu}
\maketitle                        
\thispagestyle{empty}


\noindent
{\bf Abstract -
   {\small\em In this paper, computational aspects of the \emph{panel aggregation problem} are addressed.  Motivated primarily by applications of risk assessment, an algorithm is developed for fusing large corpora of internally incoherent probability assessments.  The algorithm is characterized by a provable performance guarantee, and is demonstrated to be orders of magnitude faster than existing tools when tested on several real-world data-sets.  In addition, unexpected connections between research in risk assessment and wireless sensor networks are exposed, as several key ideas are illustrated to be useful in both fields.}
}

\vspace{0.25cm}


\vspace{-.25cm}
\section{Introduction}
\subsection{Aggregating Human Expertise}
In this paper, we address the problem of aggregating human expertise, motivated primarily by applications of risk assessment and analysis \cite{MorHen90}. In these settings, a dearth of hard data often limits one's ability to extrapolate the future from the past.   As a result, panels of human experts are frequently consulted to make forecasts about future events and to characterize the uncertainty therein. For example, stock market analysts are consulted to design risk-balanced investment portfolios, and geopolitical forecasters help construct robust policies and risk-based resource allocation schemes \cite{Lip06, WilMorKelMed06}. Typically, a multiplicity of experts are consulted in order to maximize the information available to the would-be decision-maker.  However, a panel's generally disparate opinion often needs to be fused to provide a single, coherent worldview that is useful for decision-making and analysis.

This \emph{panel aggregation problem} represents a classic example of information fusion wherein experts' forecasts must be combined for use by a centralized decision-maker.  Under various models for the information provided by the experts, the aggregation problem has been usefully addressed in fields including philosophy, law, statistics, risk analysis and computer science.  Recently, Osherson and Vardi \cite{OshVar05} considered the case where human judges provide forecasts of subjective probability for both logically simple and complex events.  A \emph{coherent approximation principle} (CAP) was proposed as a generalization of linear averaging (see, e.g., \cite{CleWin99},  \cite{Cle89}).  As discussed below, CAP is practically motivated, accommodating both incoherent (e.g., human) judges and partially specified forecasts.  However, as noted in \cite{OshVar05}, implementing CAP is NP-Hard in the general case.  Thus, for problems of interest, the CAP approach to fusion is computationally infeasible in theory and practice.  

Also in \cite{OshVar05}, Osherson and Vardi propose a method for addressing CAP's computational challenge. Termed SAPA (Simulated Annealing over Probability Arrays), their algorithm applies to a very broad class of logically complex forecasts.  Though vastly better than off-the-shelf tools, SAPA nonetheless requires many hours to aggregate forecasts provided by reasonably sized panels; CAP remains of limited use in practice. 

Nevertheless, in several experiments documented in \cite{OshVar05},  it was noted that on real-world data sets, \emph{fusing expertise using CAP (via SAPA) improves the forecasting accuracy of panel members} according to several naturally quantified measures for stochastic accuracy (we elaborate on this finding below).  This empirical result invites us to develop computationally efficient tools for implementing (or approximately implementing) CAP, so that these findings may be exploited in practice. 

Thus, the primary motivation for this paper is CAP's computational challenge.  Here, we derive a scalable algorithm for fusing forecasts of probability according to CAP.  By exploiting the logical simplicity of the events in question, a convenient application of alternating projection algorithms provides a fast tool for risk assessment with a provable performance guarantee and documented empirical success.

\subsection{Wireless Sensor Networks}
A recurrent theme in the study of wireless sensor networks (WSNs) \cite{AkySuSanCay02} is the need to exploit node-level intelligence when designing communication-efficient systems for distributed inference. With sensors that communicate inferences  (rather than raw data), future WSNs will trade computational power for energy and bandwidth.  This vision is a driver behind the demand for collaborative signal processing and for fusion strategies for aggregating inferences made by smart sensors.  As alluded to above, researchers in risk assessment have long been interested in extracting robust and calibrated forecasts from \emph{human} experts through collaboration and aggregation, and have developed a host of tools for doing so.   Thus, a secondary motivation of this paper is to connect studies in risk assessment with research in sensor networks (and vice-versa), and to expose a set of fundamental tools that may be useful for both.

\subsection{Organization}
The remainder of this paper is organized as follows.  In Section 2, we introduce notation and review alternating projection algorithms, a tool that we exploit in deriving our scalable aggregation algorithm.  In Section 3, we formalize the panel aggregation problem as an instance of information fusion, review Osherson and Vardi's coherent approximation principle, and discuss its relation to other approaches to aggregation.  In Section 4, we derive an iterative algorithm which approximately implements CAP and we discuss a theorem which characterizes the algorithm's dynamics.  In Section 5, we validate our approach with experiments on several real-world data sets.  Finally, in Section 6, we discuss extensions of the current work and connections to collaborative signal processing in WSNs.

\section{Preliminaries}
\subsection{Notation}
Let $X = (X_1, \ldots, X_n)$ be a vector of Boolean\footnote{The assumption that the variables are {B}oolean is made merely to simplify exposition; all the subsequent discussion and results hold for more general multi-valued discrete variables.} variables.  Each component of $X$ models a \textit{basic event}.  For example, the event that ``Google stock outperforms the NASDAQ in the third quarter" may be described by a Boolean variable $X_1$ whose value is $1$ if the event is true and $0$ otherwise.  $X$ therefore models a set of $n$ basic events, which could describe the performance of a set of stocks, the status of various economic indicators, the outcome of geopolitical events, etc.  

\textit{Complex events} are modeled by joining the components of $X$ with logical connectives like $\{\neg, \wedge, \vee,\ldots\}$.   For example, the complex event that ``Google stock outperforms the NASDAQ \textit{AND} the U.S. GDP increases in the third quarter" may be modeled by the conjunction $X_1 \wedge X_2$, with $X_2$ appropriately chosen.  In a slight abuse of notation, we henceforth refer to components of $X$ and logical combinations thereof as basic events and complex events, respectively.

A \textit{forecast} $(E, \hat{p})$ is an event $E$ (basic or complex) paired with a real-number $\hat{p}\in[0,1]$.  $\hat{p}$ is interpreted as an assessment of the probability that the event $E$ is true.  In the sequel, we deal with collections of forecasts $\{(E_i, \hat{p}_i)\}_{i=1}^m$, an important concept of which is \emph{probabilistic coherence}.

\begin{defn}
A set of forecasts $\{(E_i, \hat{p}_i)\}_{i=1}^m$ is \emph{probabilistically coherent} if and only if they are implied by a joint probability distribution over $X$.
\end{defn}

The following easy-to-prove lemma is important for the subsequent development.

\begin{lem}
Let $C = C(\{E_i\}_{i=1}^m) \subseteq[0,1]^m$ be the set such that $\{(E_i, \hat{p}_i)\}_{i=1}^m$ is probabilistically coherent if and only if ${\mathbf{\hat{p}}} = (\hat{p}_i)_{i=1}^m\in C$.  Then, for any set of events $\{E_i\}_{i=1}^m$, $C$ is closed and convex.
\end{lem}

\subsection{Alternating Projection Algorithms}
Let $C_1, \ldots, C_l$ be closed, convex subsets of $\bbbr^m$, whose intersection $C=\cap_{i=1}^l C_i$ is non-empty.   For any $\hat{{\mathbf{x}}}\in\bbbr^m$, let $P_C({\hat{{\mathbf{x}}}})$ denote the least-squares projection of $\hat{{\mathbf{x}}}$ onto C, i.e., 
\begin{equation}\nonumber
P_C({\hat{x}}) :=  \arg\min_{{\mathbf{x}}\in C} \|{\mathbf{x}} - \hat{\mathbf{x}}\|_2^2.
\end{equation}
Alternating projection algorithms \cite{CenZen97} provide a way to compute $P_C(\cdot)$ given $\{P_{C_i}(\cdot)\}_{i=1}^l$.  Depicted in Table 1, the von Neumann-Halperin algorithm is an example of one natural approach.
\begin{table}[htdp]
\begin{center}
\small
\begin{tabular}{|ll|}
\hline
\textbf{Initialize:} & ${\mathbf{x}}_{0}:={\mathbf{\hat{x}}}$\\
&\\
\textbf{Iterate:} & ${\mathbf{x}}_{n+1} := P_{C_{(n \mod l) + 1}}({\mathbf{x}}_n)$\\
\hline
\end{tabular}
\end{center}
\caption{The von Neumann-Halperin Algorithm} \label{vNH}
\end{table}

\noindent In words, the algorithm successively and iteratively projects onto each of the subsets.  In the case where $C_i$ is a linear subspace for all $i\in\{1,\ldots,l\}$, this algorithm was first studied by von Neumann and subsequently by Halperin.   Much of the behavior of this algorithm can be understood through Theorem \ref{vNH-thm}, the proof of which can be found in \cite{CenZen97}.  

\begin{thm}\label{vNH-thm}
Let $\{C_i\}_{i=1}^l$ be a collection of closed, convex subsets of $\bbbr^m$ whose intersection $C = \cap_{i=1}^l C_i$ is nonempty.  Let ${\mathbf{x}}_n$ be defined as in the von Neumann-Halperin algorithm. Then,  for every ${\mathbf{x}}\in C$ and every $n\geq 1$,
\begin{equation}
\nonumber\|{\mathbf{x}}_n - {\mathbf{x}}\|_2 \leq \|{\mathbf{x}}_{n-1} - {\mathbf{x}}\|_2.
\end{equation}
Moreoever, $\lim_{n\rightarrow\infty} {\mathbf{x}}_n \in \cap_{i=1}^l C_i$.  If $C_i$ is affine for all $i\in\{1,...,l\}$, then $\lim_{n\rightarrow\infty} \|{\mathbf{x}}_n - P_C(\hat{{\mathbf{x}}})\|_2 = 0$. \end{thm}

Often examined in the context of the \emph{convex feasibility problem}, the von Neumann-Halperin algorithm has been generalized in various ways to address more general convex sets and non-orthogonal projections; accordingly, the algorithm often takes on other names (e.g., Bregman's algorithm, Dykstra's algorithm). 





\section{The Panel Aggregation Problem}

\subsection{A Model}
Suppose that each of $m$ judges assesses the likelihood of a set of events; let ${\cal{E}}_i =  \{(E_{ij}, \hat{p}_{ij})\}_{j=1}^{m_i}$ denote the set of forecasts provided by judge $i$.  We assume that the events that make up ${\cal{E}}_i$ are defined over the same $X$ for all $i=1,\ldots,m$; however, we make no additional assumptions regarding the logical relationship between events in ${\cal{E}}_i$ and ${\cal{E}}_j$. In other words, we assume that panel members provide forecasts for the same ``problem domain", but may assess the likelihood for altogether different, though perhaps logically related, events.  With this model, the panel aggregation problem can be stated as follows:    

\begin{quote}\emph{Given the judges' forecasts $\{{\mathcal{E}}_i\}_{i=1}^m$, derive a coherent set of forecasts that jointly reflects the panel's expertise.}
\end{quote}

\subsection{The Coherent Approximation Principle}
Osherson and Vardi \cite{OshVar05} propose a \emph{coherent approximation principle} (CAP) for addressing the panel aggregation problem.  In particular, they suggest aggregating the panel's expertise by solving the following optimization problem:
\begin{eqnarray}
\label{CAP}\min & \sum_{i=1}^m \sum_{j=1}^{m_j} |p_{ij} - \hat{p}_{ij}|^2 &\\
\nonumber\textrm{s.t.} & \cup_{i=1}^m \{(E_{ij}, p_{ij})\}_{j=1}^{m_i} \textrm{ is coherent.}
\end{eqnarray}
Here, the optimization variables are $\{p_{ij}\}$; the events in $\{E_{ij}\}$ and the probability assessments $\{\hat{p}_{ij}\}$ are the program data.  Consistent with the definition of the panel aggregation problem in Section 3.1, the output of CAP is a coherent set of forecasts for the events in $\{E_{ij}\}$, and not (necessarily) a joint probability distribution over $X$.

By solving (\ref{CAP}), one finds the coherent forecasts that are minimally different (with respect to squared-deviation) from those provided by the panel, intuitively preserving the ``information" provided by the judges while gaining probabilistic coherence.  From a statistical perspective, computing (\ref{CAP}) can be interpreted as finding the maximum-likelihood coherent forecasts $\{p_{ij}\}$ given additive white noise corrupted observations $\{\hat{p}_{ij}\}$. Finally, CAP offers a geometric interpretation:  by Lemma 1, there exists a closed convex set $C=C(\{E_{ij}\})$ that defines the numbers which comprise coherent forecasts for the events in question; ${\mathbf{\hat{p}}}$, a vector concatenation of $\{\hat{p}_{ij}\}$, lies outside this set.  CAP suggests fusing the panel's expertise by computing the orthogonal projection of ${\mathbf{\hat{p}}}$ onto $C$.   Henceforth, the forecasts determined by solving (\ref{CAP}) will be referred to as the \emph{CAP-Aggregate} for the panel.

As discussed in \cite{OshVar05}, solving (\ref{CAP}) (and therefore, implementing CAP) is NP-Hard in the general case.  In particular, note that checking whether a set of forecasts $\{(E_{ij}, \hat{p}_{ij})\}_{j=1}^{m_i}$ is probabilistically coherent can be reduced to solving (\ref{CAP}); and checking for probabilistic coherence is strictly more general than checking whether the formulae that describe the events $\{E_{ij}\}$ are mutually satisfiable.

\subsection{Related Work}
The literature on the panel aggregation problem is expansive, as it has been touched upon in philosophy, law, statistics, risk analysis, and computer science; we refer the interested reader to the brief survey in \cite{OshVar05} for an entry point.  Here, we discuss the literature immediately relevant to CAP, and augment the survey in \cite{OshVar05} with a discussion of related work in computer science. 

Linear averaging \cite{CleWin99},\cite{Cle89} is arguably the most popular aggregation principle, given its simplicity, various axiomatic justifications, and documented empirical success.   To illustrate this natural approach, consider the panel exhibited in Table 2.  Here, three judges provide forecasts for three events, a conjunction and its conjuncts.  The ``Aggregate" forecast is the simple un-weighted average of the three judges' forecasts. Though appealing, linear averaging is not without pitfalls, as can be illustrated with a few examples. 

For instance, an underlying assumption in linear averaging is that each judge is probabilistically coherent.  Averaging is appropriate under this assumption since (by Lemma 1) the linear averaged aggregate is probabilistically coherent whenever the individual judges are coherent.  
However, in applications of interest, the judges are humans, who are notoriously incoherent.  For example, the \emph{conjunction fallacy}, a robust finding from psychology \cite{KahTve00}, \cite{TenBonOsh04}, demonstrates that human judges (even experts!) often assign higher probability to a conjunction that its conjuncts.  Table 2 illustrates such a case. In particular, note that ``Chris" is incoherent since the probability assigned to the event $p\wedge q$ is greater than the probability assigned to $q$, i.e., $0.6 > 0.0$; the linear averaged aggregate is similarly incoherent. Thus, though linear averaging naturally addresses \emph{inter-judge disagreement}, it will not in general provide a coherent aggregate when individual judges are themselves incoherent.  

A clever analyst may circumvent this problem by soliciting forecasts for logically independent events.  Such a strategy may work in isolated cases, but it is not a general solution and may ultimately require the analyst to ignore subtleties in the experts' forecasts. For example, a market analyst may complement a forecast concerning the NASDAQ by forecasting a correlation between the NASDAQ and currency exchanges; a geopolitical expert may assess the likelihood of a terror attack in a \emph{particular} city and also by forecast the probability of an attack in \emph{any} city.  In short, there is information to be gleaned from forecasts for logically complex events:  practical aggregation principles should recognize this fact while accommodating intra-judge incoherence.
\begin{table}[htdp]
\begin{center}
\small
\begin{tabular}{|c|c|c|c|c|}
\hline 
& Alice & Bob & Chris & Aggregate \\
\hline
$p$ & 0.75 & 0.60 & 0.95 &  0.67\\
\hline
$q$ &  0.20 & 0.10 & \textbf{0.00} & \textbf{0.10} \\
\hline
$p\wedge q$  & 0.10 & 0.10 & \textbf{0.60} & \textbf{0.20}\\
\hline
\end{tabular}
\end{center}
\label{default}
\caption{Linear Averaging: Incoherent Judges}
\end{table}%

In practice, human judges may be unable or unwilling to offer forecasts for every event in question.  Communication constraints may preclude judges from collaborating, or individual judges may find themselves unqualified to forecast the likelihood of particular events.  To reuse an earlier example, an analyst may be unwilling to forecast events pertaining to the technology sector but may willing do discuss the correlation between the NASDAQ and the currency exchanges.  Such a case is illustrated in Table 3, where each judge provides an incomplete but coherent set of forecasts.  The incoherence of the pairwise average aggregate demonstrates that linear averaging is also inappropriate in the case where judges provide only partial forecasts.

Given the aforementioned limitations of linear averaging, a natural question arises:  how should one aggregate (i.e., fuse) the opinion expressed by incoherent judges on overlapping but generally different sets of logically complex events? CAP addresses this question by generalizing linear averaging.  In particular, note that the CAP aggregate \emph{equals} the un-weighted averaged aggregate whenever probabilistically coherent judges provide forecasts for the same set of events. 

\begin{table}[htdp]
\begin{center}
\small
\begin{tabular}{|c|c|c|c|c|}
\hline 
& Alice & Bob & Chris & Aggregate \\
\hline
$p$ & 0.75 & 0.60 &\textbf{NA} &  0.67\\
\hline
$q$ & 0.20 & \textbf{NA} & 0.00 & \textbf{0.10} \\
\hline
$p\wedge q$  & \textbf{NA} & 0.40 &0.00  & \textbf{0.20}\\
\hline
\end{tabular}
\end{center}
\label{default}
\caption{Linear Averaging: Partial Forecasts}
\end{table}%

Lindley et al. \cite{LinTveBro79} consider a Bayesian approach to reconciling probability forecasts, whereby ``noisy" observations $\{\hat{p}_{ij}\}$  are assumed to arise from a coherent set $\{p_{ij}\}$.  CAP can be viewed as a special-case of their model, since as discussed above, the solution to (\ref{CAP}) admits a Bayesian interpretation as the maximum-likelihood coherent forecasts given additive white noise corrupted observations $\{\hat{p}_{ij}\}$.  However, note that \cite{LinTveBro79} sought to eliminate incoherence from a single judge, whereas CAP was introduced to address the panel aggregation problem.  Moreover, Osherson and Vardi were motivated by non-statistical interpretations of CAP and as here, addressed the computational issue of implementing CAP.

A panel-aggregation problem is addressed in the ``online" learning model, which is frequently studied in learning theory \cite{CesFreHauHelSchWar97} \cite{FreSch97b}. In that setting, a panel of experts predicts the true outcome of a set of events.  A central agent constructs its own forecast by fusing the experts' predictions, and upon learning the truth, suffers a loss sometimes specified by a quadratic penalty function.  In repeated trials, the agent updates its fusion rule (e.g., the ``weights" in a weighted average), taking into account the performance of each expert.  Under minimal assumptions on the evolution of these trials, bounds are derived that compare the trial-averaged performance of the central agent with that of the best (weighted combination of) expert(s).  In contrast to the current framework, the online model typically assumes that each expert provides a forecast for the same event or partition of events.  Thus, fusion strategies such as weighted averaging are appropriate in the online model, for the same reasons discussed above.  Also, observe that the present model concerns a single ``trial", not many. 

Finally, proponents of Dempster-Shafer theory \cite{Sha76} (and associated fusion rules) object to probability as an idiom for belief, in part because of its inability to distinguish uncertainty from ignorance.  The merits of Dempster-Shafer aside, one could argue for abstention as an expression of ignorance.  As the preceding examples illustrate, even abstaining experts may disagree (i.e., experts' forecasts may be mutually incoherent), and therefore the panel aggregation problem remains.  Thus, CAP is a natural aggregation principle in the setting where judges express uncertainty with probability and ignorance through abstention, and thereby extends the utility of probabilistic forecasts by affording experts more expressive beliefs with abstention.

\section{A Scalable Approach}
In principle, implementing CAP by solving (\ref{CAP}) can be accomplished using quadratic programming.  In the general case, this approach requires a representation of joint distributions on $X$, for which $O(2^n)$ free variables are necessary. For panels that assess relatively small numbers of events, the quadratic programming approach is nonetheless feasible.  In cases of interest, hundreds of judges forecast thousands of events, yet off-the-shelf tools for solving quadratic programs do not scale.

Nevertheless, the logical complexity of the events assessed by human judges is usually bounded.  For example, experts are often constrained to forecast events with no more than $three$ literals (e.g., three-term conjunctions).  The idea at the heart of our approach is to exploit such logical simplicity by decomposing (\ref{CAP}) into a collection of small sub-problems, each of which can be solved quickly using off-the-shelf tools.  

We now present our main result, a general algorithm for aggregating large corpora of probability forecasts.  To aid exposition, let us do away with the multi-judge distinction by assuming that there is a single body of forecasts ${\cal{E}} = \{(E_i, \hat{p}_i)\}_{i=1}^m$. We do so without loss of generality, since we may construct ${\cal{E}}$ by pooling all the judges' forecasts into a single set.  Also, let us assume that every event in $\{E_i\}_{i=1}^m$ is unique.  Below, we demonstrate how this assumption may be relaxed.
\subsection{A General Algorithm}
To state our general algorithm, it is helpful to introduce a notion of \emph{local coherence}.  Let $\{(E_i,\hat{p}_i)\}_{i=1}^m$ be a collection of forecasts and let $\sigma\subseteq\{1,\ldots,m\}$.  The requirement that $\{(E_i,\hat{p}_i)\}_{i=1}^m$ be probabilistically coherent can be relaxed by requiring only the subset $\{(E_i,\hat{p}_i)\}_{i\in\sigma}$ be coherent.  For notational convenience, we henceforth say that  $\{(E_i,\hat{p}_i)\}_{i=1}^m$ is \emph{locally coherent with respect to $\sigma$} whenever  $\{(E_i,\hat{p}_i)\}_{i\in\sigma}$ is coherent.

With this formalism, note that ``global" coherence is recovered by taking $\sigma=\{1,\ldots,m\}$. Moreover, note that any probabilistically coherent set $\{(E_i,\hat{p}_i)\}_{i=1}^m$ must be locally coherent with respect to $\sigma$ for all $\sigma\subseteq\{1,\ldots, m\}$.   

With that, let us relax (\ref{CAP}) by choosing a collection of subsets $\{\sigma_j\}_{j=1}^l$ and defining the following optimization problem.
\begin{eqnarray}
\label{CAP-relax}\min & \sum_{i=1}^{m} |p_{i} - \hat{p_{i}}|^2 &\\
\nonumber\textrm{s.t.} & \{(E_{i}, p_{i})\}_{i\in\sigma_j} \textrm{is coherent} &\forall j=1,\ldots, l
\end{eqnarray}
To emphasize,  (\ref{CAP-relax}) is a relaxation of (\ref{CAP}), since in general local coherence does not imply global coherence.  However, this relaxation permits a geometric interpretation, as a projection onto the intersection of $l$ convex sets.  Thus, alternating projection algorithms are applicable to solving (\ref{CAP-relax}).  In particular, an algorithm for solving (\ref{CAP-relax}) is detailed in Table 4; note that it is exactly the von Neumann-Halperin algorithm interpreted in the language of the panel aggregation problem.  

\begin{table}[htdp]
\begin{center}
\small
\begin{tabular}{|ll|}
\hline
\textbf{Input:} & $\{(E_i, \hat{p}_i)\}_{i=1}^m$\\
&\\
\textbf{Initalize:} &  Auxiliary forecasts $\{(E_i, q_i)\}_{i=1}^m$, with $q_i := \hat{p}_i$.\\
&\\
\textbf{Step 1:} & Design $\{\sigma_j\}_{j=1}^l$ with $\sigma_j\subseteq\{1,\ldots, m\}$.\\
&\\
\textbf{Step 2:} & for $t=1, \ldots, T$\\ 
& \,\,\,\,for $j=1,\ldots, l$ \\
& \hspace{7mm}${\mathbf{p}}_{tj} := \arg \min \sum_{i=1}^m |p_i - q_i|^2$\\
& \hspace{24mm}s.t. $\{(E_i, p_i)\}_{i\in\sigma_j}$ is coherent. \\
& \hspace{7mm}Update $\{(E_i, q_i)\}_{i\in\sigma_j}\leftarrow \{(E_i, p_{tj, i})\}_{i\in\sigma_j}$\\
\textbf{Output:}& $\{(E_i, q_i)\}_{i=1}^m$.\\
\hline
\end{tabular}
\end{center}
\caption{A Scalable Approach to Aggregation} \label{vNH}
\end{table}

In this algorithm, computation occurs in the inner loop, when projecting ${\mathbf{q}}$ onto a set of local coherence constraints.  This computation requires only $|\sigma_j|$ forecasts, since $p_{tj, i} = q_i$ for all $i\notin \sigma_j$, and can be achieved using off-the-self tools for quadratic programming or more specialized tools like SAPA.

The crucial step in this algorithm is Step 1, designing $\{\sigma_j\}_{j=1}^l$.  Intuitively, the fewer events that each subset contains, the faster each inner computation can run.  However, as subsets get larger, a richer set of coherence constraints are represented and thus, the solution to (\ref{CAP-relax}) more closely approximates the CAP-aggregate.  When designing $\{\sigma_j\}_{j=1}^l$, one must therefore strike a balance between \emph{approximation} and \emph{speed}.  

A natural way to make this trade-off is by exploiting the logical simplicity of the events in question. To illustrate, consider the case where the events in $\{E_i\}_{i=1}^m$ are constrained to be basic events, negations of basic events, and two-term conjunctions (or disjunctions) of the basic events or their negations.  A sample set of events that meet these criteria are drawn in Figures 1, 2, and 3 (ignoring the dashed lines for a moment).

The linear averaging approach to aggregation can be viewed as a special case of this general method, where one subset is chosen per event; these subsets are depicted by the dashed lines in Figure 1.  This highly local approach can be implemented very quickly, however the solution to (\ref{CAP-relax}) may poorly approximate the CAP-aggregate since few of the coherence constraints are represented.  CAP, on the other hand, groups all the events into a single subset, requiring global coherence; this case is depicted in Figure 2.  The CAP approach represents all the coherence constraints, but as discussed above, is computationally infeasible in practice.

A cleverer design may select subsets according to the logical relationship between the events in question.  In Figure 3, for example, it is proposed to group basic events with their negations, and conjunction (disjunctions) with their corresponding conjuncts (disjuncts).  By choosing \emph{all} subsets of this form, we enforce a very strong set of local coherence constraints; crucially, however, each subset contains at most three events.  Intuitively, solving (\ref{CAP-relax}) using these subsets will quickly approximate the CAP-aggregate given the balance we have struck between approximation and speed.  This intuition is borne out in the experiments.

\begin{figure}[htb]
\begin{center}
\includegraphics[width=3.0in]{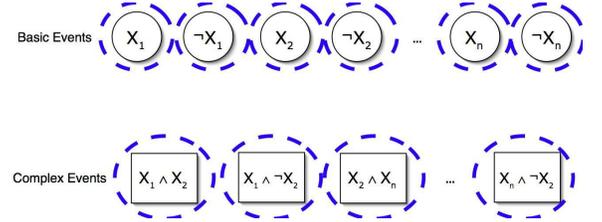}
\end{center}
\vspace{-.5cm}
\caption{Linear Averaging}
\end{figure}
    
\begin{figure}[htb]
\begin{center}
\includegraphics[width=3.0in]{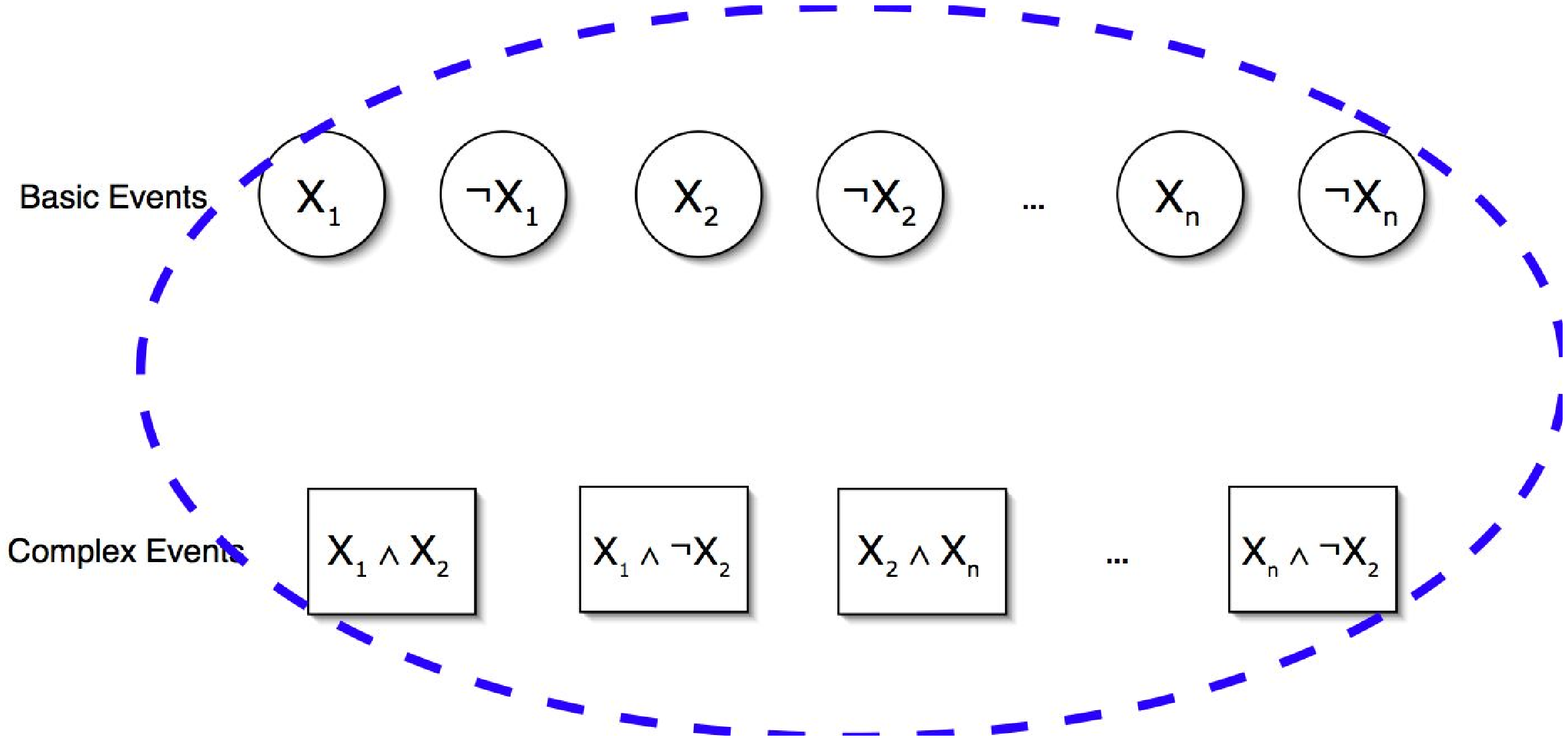}
\end{center}
\vspace{-.5cm}
\caption{CAP}
\end{figure}

\begin{figure}[htb]
\begin{center}
\includegraphics[width=3.0in]{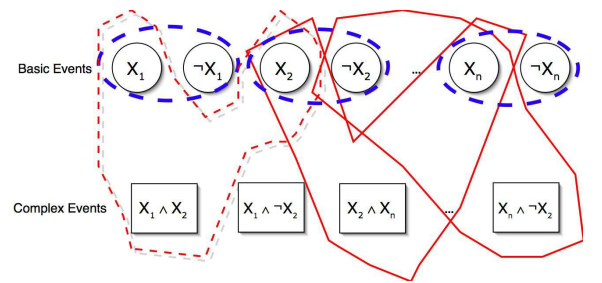}
\end{center}
\vspace{-.5cm}
\caption{A Scalable Approach}
\vspace{-.5cm}
\end{figure}

\subsection{Comments}
First, let us emphasize that $\{\sigma_j\}_{j=1}^l$ is a design parameter. Depending on the design, the output forecasts may or may not be coherent;  recall, the algorithm solves (\ref{CAP-relax}), a \emph{relaxation} of CAP.  Intuitively, however, for any $\{\sigma_j\}_{j=1}^l$ the output will be closer to coherence, since it will satisfy a set of local coherence constraints.  This intuition is formalized by Theorem 2 below.  

Second, in Step 2, the local coherence constraints are addressed in sequence.  Note that this ordering is non-essential and parallelism may be introduced. In particular, two projections can occur simultaneously as long as there is no overlap between the events in question (i.e., two projections cannot change the same variables simultaneously).  


Next, the assumption that each event in $\{E_i\}_{i=1}^m$ is unique can be relaxed.  For any set $\{(E_i, \hat{p}_i)\}_{i=1}^m$ with $N_j$ forecasts for each unique event $F_j\in\{E_i\}_{i=1}^m$, one can construct a new set $\{(F_j, \hat{q}_j)\}$ with $\hat{q}_j = \frac{1}{N_j} \sum_{i: E_i = F_j} \hat{p}_i$.  Then, solving 
\begin{eqnarray}
\nonumber \min & \sum_{j} N_j |p_j - \hat{q}_j|^2\\
\nonumber\textrm{s.t.} &  \{(F_{j}, p_{j})\} \textrm{ is coherent.}
\end{eqnarray}
is equivalent to solving (\ref{CAP}) with $\{(E_i, \hat{p}_i)\}_{i=1}^m$.  The same trick can be applied to the relaxation (\ref{CAP-relax}), and the algorithm in Table 4 can be adjusted similarly.

Finally, the algorithm depicted in Table 4 permits a performance guarantee.  In particular, assume that after learning the ``truth" of the events in question, the accuracy of the forecasts in $\{(E_i, \hat{p}_i)\}_{i=1}^m$ is assessed using the \emph{Brier score} \cite{Bri50}, a quadratic penalty:
\begin{equation}
\label{QP} QP(\{(E_i, \hat{p}_i)\}) = \sum_{i: E_i = \textrm{TRUE}} (1 - \hat{p}_i)^2 + \sum_{i: E_i = \textrm{FALSE}} (0 - \hat{p}_i)^2
\end{equation}

The algorithm in Table 4 offers a stepwise improvement in accuracy as measured by the Brier score, independent of the truth or falsity of the events in question.  Theorem 2 formalizes this important fact.

\begin{thm}
Let $\{(E_i, q_{T,i})\}_{i=1}^m$ denote the set of forecasts output by the algorithm after running $T$ iterations with input forecasts $\{(E_i, \hat{p}_{i})\}_{i=1}^m$.  Then,
\begin{equation}
\nonumber QP(\{(E_i, q_{T, i})\}) \leq QP(\{(E_i, q_{T-1, i})\})
\end{equation}
under every realizable truth assignment to the events in $\{E_i\}_{i=1}^m$.  Moreover, as $T\rightarrow\infty$, the output forecasts converge (i.e., ${\mathbf{q}}_{T}$ converges in norm), and are locally coherent with respect to $\sigma_j$ for all $j=1,\ldots, l$.
\end{thm}
The proof of Theorem 2 follows from Theorem 1 and de Finetti's Theorem \cite{Def74}, \cite{PreOshKulPoo05a}.  If $\{(E_i, \hat{p}_i)\}_{i=1}^m$ contains a single judge's forecasts (i.e., the algorithm is applied to eliminate intra-judge incoherence), then Theorem 2 predicts a step-wise improvement in the accuracy of that judge.  If instead $\{(E_i, \hat{p}_i)\}_{i=1}^m$ contains a panel's forecasts, then Theorem 2 predicts that at each step, a randomly selected judge will improve on average.  

Note that $T$, the number of iterations through the forecasts, is a second design parameter for this algorithm that in principle must be tuned.  However, Theorem 2 demonstrates a sense in which performance is monotonic in $T$.  Moreover, for any $T$, the output forecasts will be more accurate than the input forecasts (with respect to the Brier score), independent of the truth or falsity of the events in question.

\section{Experiments}
In this section, we empirically validate the aggregation algorithm presented in Section 4.  In particular, our experiments focus on two issues: (i) the effect that aggregation (i.e., fusion) has on the panel's forecasting accuracy and (ii) how the algorithm scales to large data sets, i.e., how ``fast" the algorithm is in practice.
\subsection{The Data}
Five previously collected data sets will be used in these experiments.  The STCK database was first published in \cite{OshVar05} and contains forecasts made by MBA students at Rice University on events pertaining to $10$ stocks in the third quarter of 2000; the FIN database is documented in \cite{BatBreOshVarTsa02} and summarizes forecasts made by students at Rice on events related to various economic indicators in the fourth quarter of 2001; the NBA1 and NBA2 data sets appeared in \cite{BatBreOshVarTsa02} and detail forecasts made by self-proclaimed basketball enthusiasts regarding the outcome of two Houston Rockets National Basketball Association games; the HSTN data set \cite{HenHarOsh05} contains forecasts made by Houston homeowners on events pertaining to the local real-estate market and pollution.

In each of the five data sets, subjects were asked to assess the likelihood of $34$ randomly selected basic ($10$) and complex ($24$) events.  The complex events were constrained to have one the following forms: $p \wedge q$, $p\wedge \neg q$, $p \vee q$, or $p \vee \neg q$.  The number of subjects (i.e., the size of the panel) per data set is summarized in Table 5, as is the total number of basic events (i.e., the length of $X$) from which the forecasted events were constructed. Due to the random allocation of events per subject, multiple experts often provided forecasts pertaining to the same event.  In Table 5, ``Events/Agg" describes the number of unique events per panel.

\begin{table}[htdp]
\begin{center}
\small
\begin{tabular}{|c|c|c|c|c|c|}
\hline
& \textbf{STCK} & FIN & NBA1 & NBA2 & HSTN \\
\hline
Subjects &47 &31 & 29&36 & 17 \\
\hline
Basic Events & \textbf{30}&10& 10&10 &10 \\
\hline
Events/Agg. & \textbf{1598}& 1054& 986&1224 & 578\\
\hline
\end{tabular}
\end{center}
\label{default}
\caption{Data Summary}
\vspace{-.5cm}
\end{table}%
\subsection{The Method}
In each of the following experiments, we employ the aggregation algorithm detailed in Section IV.  Since in each data set, complex events are constrained to one of the forms $p \wedge q$, $p\wedge \neg q$, $p \vee q$, or $p \vee \neg q$, subsets are chosen precisely as illustrated by Figure 3.  Interestingly, for these subsets, deterministic rules can be derived for solving each optimization in Step 2 (Table 4); we forego describing these easily derived rules in the interest of space.

For every forecast reported in each database, the truth-value of the corresponding event is known.  This allows us to assess the accuracy of various forecasts \textit{a posteriori}.   Here, accuracy is measured using the Brier score (\ref{QP}) and slope, which is defined as follows: if $m_T$ denotes the number of true events in $\{E_i\}_{i=1}^m$, slope measures the stochastic accuracy of the forecasts $\{(E_i, \hat{p}_i)\}_{i=1}^m$ using $\frac{1}{m_T} \sum_{i: E_i = \textrm{TRUE}} \hat{p}_i - \frac{1}{m-m_T}  \sum_{i: E_i = \textrm{FALSE}}\hat{p}_i$; higher slope indicates more accurate forecasts.  We assess the accuracy of forecasts in four cases of interest.

\begin{itemize}
\item \emph{Raw:} the accuracy of the judge's raw forecasts. The average accuracy of each judge's unprocessed forecasts is reported.

\item \emph{Individual:}  the accuracy after eliminating intra-judge incoherence (i.e., after running the algorithm on each individual judge). The average accuracy of the judge's forecasts after processing is reported.

\item \emph{Aggregate:} the accuracy after aggregation using our method.  The accuracy of each judge is assessed after replacing her original forecasts with the aggregate forecasts (for the same events);  the average judge's score is reported.

\item \emph{Linear Avg.:} the accuracy of the linear averaged aggregate. The accuracy of each judge is assessed after replacing her original forecasts with the linear averaged aggregate (again, for the same events); the average judge's score is reported.
\end{itemize}


\noindent Note that when measuring accuracy with slope, the score reported for \emph{linear averaging} will be the same as that which is reported for \emph{raw}.

\subsection{Experiment 1: Scalability}
Figures 4 and 5 detail the average Brier score achieved by the panel vs.\ the number of iterations ($T$) made by our algorithm, in the \emph{Individual} and \emph{Aggregation} cases respectively.  Note that the monotonicity of these plots is predicted by Theorem 2.  In both cases and in every data set, the algorithm converges within $10$ iterations through the forecasts.

From a computational perspective, the most interesting data set is the STCK database, since it contains the largest number of unique events per aggregate and the most basic events.  On a 1GHz PowerPC G4, aggregating the database of 1598 forecasts took approximately 10s. In contrast, the rival method SAPA \cite{OshVar05} was reported to take multiple hours.  Incidentally, the time required to eliminate incoherence from individual judges was less than 0.6s.  
\begin{figure}[htb]
\begin{center}
\includegraphics[width=3.1in]{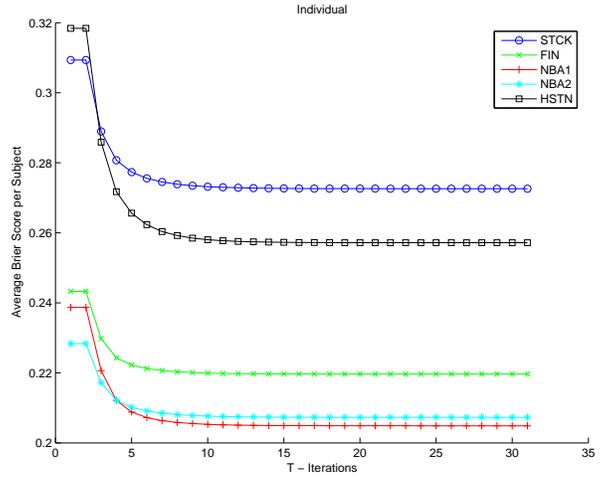}
\end{center}
\vspace{-3mm}
\caption{\emph{Individual}: Average Brier Score vs. $T$.}
\vspace{-.2cm}
\end{figure}
\begin{figure}[htb]
\begin{center}
\includegraphics[width=3.1in]{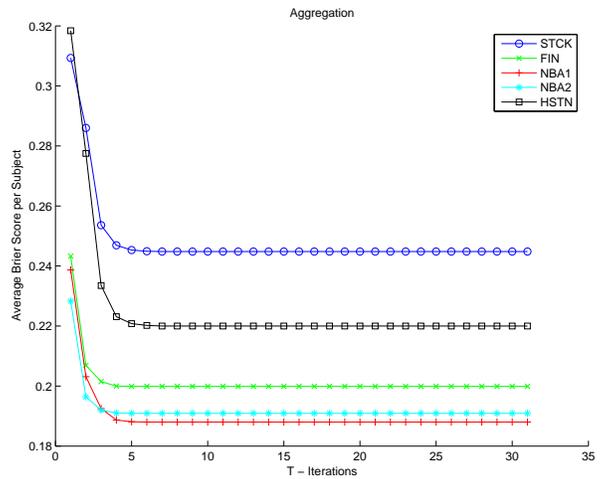}
\end{center}
\vspace{-3mm}
\caption{\emph{Aggregate}: Average Brier Score vs. $T$.}
\vspace{-.3cm}
\end{figure}

\subsection{Experiment 2: Forecasting Accuracy}
Osherson and Vardi \cite{OshVar05} report three important empirical findings.  First, they observe that eliminating intra-judge incoherence improves the forecasting accuracy of individual judges (i.e., \emph{Individual} is better than \emph{Raw}).  Second, they observe that panel aggregation improves the forecasting accuracy of panel members (i.e., \emph{Aggregate} improves over \emph{Raw}).  Finally, \cite{OshVar05} reports that aggregation improves the accuracy of panel members as compared to incoherence-corrected forecasts (i.e., \emph{Aggregate} improves over \emph{Indivdual}).  Discussed in part in reference \cite{OshVar05}, these findings are anticipated by de Finetti's theorem \cite{Def74} when accuracy is assessed using the Brier score.  However, Osherson and Vardi's findings hold up under alternative accuracy measurements including slope. 

In the previous section, we documented a several orders of magnitude speed-up.  Here, we question whether this has been achieved at the expense of accuracy. In particular, we question whether Osherson and Vardi's empirical observations hold up when using our method.  Tables 6 and 7 summarize the result for Brier score and slope, respectively.  These results are in agreement with the findings of Osherson and Vardi except that the aggregate slopes are not consistently higher than for the individual application of our algorithm.  As reported in reference \cite{OshVar05} for the STCK dataset, the SAPA method yielded average per subject accuracy as $0.276$ (\emph{Indivdual}), while the ``optimal" CAP calculation computed using quadratic programming yieled $0.272$ (\emph{Individual}).  Note that CAP We thus conclude that the proposed method provides a significant computational speed-up while achieving competitive forecasting gains.
 \begin{table}[htdp]
\begin{center}
\footnotesize
\begin{tabular}{|c|c|c|c|c|c|}
\hline
& STCK & FIN & NBA1 & NBA2 & HSTN \\
\hline
Raw & 0.309 & 0.243 &0.239 & 0.228 & 0.318\\
\hline
Individual &0.273 & 0.220&0.205 & 0.207& 0.257\\
\hline
Aggregate &0.245 & 0.200& 0.188& 0.191& 0.220\\
\hline
Linear Avg. &0.286 & 0.207 & 0.203& 0.196& 0.234\\
\hline
\end{tabular}
\end{center}
\label{default}
\caption{Forecasting Accuracy: Brier Score}
\end{table}%

\begin{table}[htdp]
\begin{center}
\footnotesize
\begin{tabular}{|c|c|c|c|c|c|}
\hline
& STCK & FIN & NBA1 & NBA2 & HSTN \\
\hline
Raw & 0.064& 0.153& 0.140& 0.141& 0.129\\
\hline
Individual &0.109 & 0.172 & 0.186 & 0.169 & 0.210\\
\hline
Aggregate & 0.114 & 0.153& 0.173& 0.150& 0.202\\
\hline
\end{tabular}
\end{center}
\label{default}
\caption{Forecasting Accuracy: Slope}
\end{table}%
\section{Discussion}
An underlying assumption of the current study is that the Brier-score (e.g., squared-error) is the appropriate measure for assessing forecasting accuracy and probabilistic (in)coherence.  However, de Finetti's theorem, the von-Neumann-Halperin algorithm (and generalizations such as Dykstra's algorithm) have all been extended to a wide class of distance measures known as \emph{Bregman divergences} \cite{Bre67} (which include the Brier-score and relative-entropy as special cases); for details, see \cite{CenZen97} and \cite{PreOshKulPoo05a}.  As a result, our methods and analysis can be generalized to accommodate a large class of alternative accuracy measurements.


The message-passing algorithm derived in Section 4 is reminiscent of belief propagation, the sum-product algorithm, and junction-trees more generally\footnote{The authors thank David Blei for helpful discussions in regards to graphical models.}.  It is thus natural to ask (i) whether CAP could be solved using an appropriate factor graph representation and the junction tree algorithm and (ii) whether the algorithm derived in Section 4 can be viewed as an instantiation of one such approach.  Addressing (ii) may require one to interpret alternating projection algorithms in the context of the junction-tree algorithm applied to a factor graph representation of our local coherence constraints. 

Since researchers in wireless sensor networks are interested in similar aggregation problems, it is natural to ask whether these tools are applicable in a WSN setting where the ``experts" are electro-mechanical sensors.  If in a given WSN application, sensors provide forecasts of probability for both logically simple and complex events, then these tools are immediately applicable.  However, the general idea of relaxing a projection by exploiting an underlying notion of locality is more widely applicable.  For example, in \cite{PreKulPoo06a}, a distributed algorithm is constructed for collaboratively training least-square kernel regression estimators.  Similarly to above, the algorithm was derived using alternating projection algorithms applied to network topology dependent relaxation of the classical least-squares estimator.
\vspace{-5mm}
\small
\bibliography{../../../Aggregation}
\bibliographystyle{plain}
\end{document}